\documentclass[10pt]{article}
\usepackage{arxiv}

\usepackage[T1]{fontenc}
\usepackage[utf8]{inputenc}
\usepackage{textcomp}
\usepackage{amsfonts,amssymb,amsmath}

\usepackage{graphics,graphicx}
\usepackage{booktabs}
\usepackage{tabularx}
\usepackage{array}
\usepackage{color}
\usepackage{float}
\usepackage{caption}
\usepackage{placeins}
\usepackage{enumerate}
\usepackage{appendix}
\usepackage{amsthm}
\usepackage{multirow}
\usepackage{adjustbox}
\usepackage{comment}

\usepackage{natbib}

\usepackage[hyphens]{url}
\usepackage{hyperref}
\hypersetup{
    colorlinks=true,
    linkcolor=blue,
    filecolor=magenta,
    urlcolor=cyan,
    citecolor=blue,
}

\input{glyphtounicode}
\newcommand{\tablefont}{\small}
\newenvironment{tabnote}
  {\par\smallskip\noindent\begin{minipage}{\linewidth}\footnotesize}
  {\end{minipage}\par}
\newcommand{\tbl}[3]{%
  #1%
  \begin{center}
    \begin{adjustbox}{max width=\textwidth}
      #2%
    \end{adjustbox}
  \end{center}
  #3%
}
\newcommand{\email}[1]{\texttt{#1}}

\title{CHiPS: Character Histograms and Positional Signals for Lightweight Authorship Attribution in Romanian Texts}

\author{
  Sanda-Maria Avram\textsuperscript{1}\\
  \normalfont\textsuperscript{1}Department of Computer Science,\\
  Faculty of Mathematics and Computer Science,\\
  Babe{\c{s}}-Bolyai University,\\
  Cluj-Napoca, Romania\\
  Email: \email{sanda.avram@ubbcluj.ro}
  \And
  George C. \c{T}urca\c{s}\textsuperscript{2}\\
  \normalfont\textsuperscript{2}Department of Mathematics,\\
  Faculty of Mathematics and Computer Science,\\
  Babe{\c{s}}-Bolyai University,\\
  Cluj-Napoca, Romania\\
  Email:
  \email{george.turcas@ubbcluj.ro}\\
}

\begin{document}

\maketitle

\begin{abstract}
We propose CHiPS, a lightweight character-level authorship attribution method for Romanian texts. All reported experiments are closed-set: the true author is one of the candidate authors in the training data. CHiPS studies two complementary fingerprints of writing style: CH-SVM, a character-histogram classifier based on one-character marginal distributions, and FFT12-LR, a positional-signal classifier that represents selected characters and punctuation classes as impulse trains (binary indicator sequences over character positions) and extracts Fourier/Welch spectral descriptors. We also report CHiPS-F, a leakage-safe decision-level fusion variant, and an optional top-5 listwise reranker trained only on out-of-fold predictions. The method requires no tokenization, syntactic analysis, pretrained language model, or transformer fine-tuning, and it avoids character $n$-gram features with $n \geq 2$ in the histogram component. On a locked grouped ROST split comprising 400 files from 392 source-text groups, written by 10 authors, with source-text-level evaluation and grouped five-fold model selection, CHiPS-F reaches 0.9310 accuracy and 0.9341 macro-F1. A matched but unrestricted character 2--5-gram TF--IDF SVM comparator reaches 1.0000 accuracy and macro-F1 on the same held-out groups, so the contribution is not a claim of best possible classification accuracy. Instead, the experiments ask how far restricted, transparent character evidence can go under strict leakage control. On ROSTories-cleaned, a secondary ROST-overlapping corpus comprising 1,248 files from 1,240 source-text groups, written by 19 authors, the same protocol gives 0.8919 accuracy and 0.8708 macro-F1 for CHiPS-R.

\end{abstract}


\section{Introduction}

Authorship attribution is a long-standing NLP and computational stylometry problem. Given a text of unknown or disputed origin, the task is to identify its most likely author from a set of candidates. This paper studies the closed-set setting, where the correct author is assumed to be among the candidate authors. The problem remains practically relevant in digital humanities, forensic linguistics, plagiarism investigation, literary history, and the organization of digital text collections. For Romanian, relative to English, there are fewer standardized public benchmarks and NLP resources, while genre and period variation can be substantial. This motivates attribution methods that can be trained and inspected without depending on a large language-specific preprocessing pipeline, as explored in \cite{stamatatos2009survey,koppel2009computational,he2024authorshipsurvey,avram2022comparison,avram2025bertrost,nitu2024lessresourced}. Character-level attribution has a useful precedent here: early work showed that character $n$-gram models can support language-independent authorship attribution with little language-specific preprocessing \cite{peng2003language}.

Recent work on Romanian authorship attribution has explored both traditional machine-learning features and transformer-based models \cite{avram2022comparison,avram2025bertrost,nitu2024lessresourced}. These approaches are valuable, but they also raise a methodological question:

\begin{center}
\textit{How much authorial signal is already present in character-level evidence?}
\end{center}

This paper introduces CHiPS, a family of character-level classifiers built around global character-frequency fingerprints and positional character signals. The method does not rely on word segmentation, part-of-speech tagging, syntactic parsing, contextual embeddings, pretrained language models, or neural fine-tuning. This simplicity makes it transparent, reproducible, and portable across other low-resource languages.

The central hypothesis is that authors might leave stable traces in the distribution and placement of characters. These traces include concrete habits in spelling, punctuation, diacritic use, spacing, casing, and character flow. Some may also reflect rhythm, sentence construction, lexical preferences visible at the character level, or editorial conventions. Character-level evidence is attractive because it sits below explicit topic choice and does not require linguistic annotation. One can argue that vocabulary is often conscious.
Character usage patterns are largely unconscious \cite{stamatatos2009survey}. Unlike lexical choices, which are often more topic-dependent and more susceptible to deliberate manipulation, character distributions reflect stable stylistic habits related to orthography, punctuation, spacing, morphology, and rhythmic text organization \cite{sapkota2015character,stamatatos2009survey}. Unlike character $n$-gram methods, CHiPS does not rely on contiguous character subsequences. Its histogram component uses only one-character marginal frequencies, while its positional component treats selected characters and character classes as sparse signals over text positions. All features are derived from raw characters after normalization.

The contributions of this paper are:
\begin{enumerate}
  \item[(i)] a leakage-aware character-level evaluation protocol for Romanian authorship attribution, in which all fragments from the same source text are kept together across training, validation, and held-out test partitions;
  \item[(ii)] CH-SVM, a transparent character-histogram classifier using only one-character marginal frequencies and a small number of global character statistics;
  \item[(iii)] FFT12-LR, a positional-signal classifier that maps selected character classes to impulse trains and summarizes them using Fourier/Welch spectral descriptors;
  \item[(iv)] CHiPS-F, a decision-level fusion strategy that combines distributional and positional character evidence without using word features, embeddings, or character $n$-grams of length $n \geq 2$ in the histogram component;
  \item[(v)] CHiPS-R, a leakage-safe top-5 listwise reranking layer that only reranks plausible candidate authors using out-of-fold train/validation base-model predictions and is reported as an ablation unless it improves held-out behavior under the prespecified protocol;
  \item[(vi)] a locked ROST evaluation with source-text grouping, grouped five-fold model selection on train/validation data, and held-out test reporting, plus a secondary ROSTories-cleaned evaluation that is kept separate because the broader corpus includes overlapping ROST material.
\end{enumerate}

\section{Related Work}

\subsection{Authorship attribution and stylometry}

Authorship attribution has a long tradition in computational stylometry, with methods ranging from manually selected style markers to statistical and machine-learning classifiers. Surveys by \cite{stamatatos2009survey} and \cite{koppel2009computational} describe the main problem settings, feature families, and evaluation concerns, including the distinction between closed-set and open-set attribution. More recent surveys such as \cite{he2024authorshipsurvey} revisit these issues in the context of contemporary neural models and increasingly diverse digital text collections.

For the present work, the important point is methodological rather than historical: strong attribution systems often depend on features that are easy to extract and difficult for authors to control consciously. Character distributions are a natural example. They sit below word choice and syntax, yet they can encode punctuation habits, spelling conventions, diacritic patterns, spacing, and recurring fragments of morphology.

\subsection{Character-level methods}

Character-level approaches are well established in authorship attribution. \cite{peng2003language} showed that character-level $n$-gram language models can support language-independent authorship attribution, while \cite{sapkota2015character} demonstrated that different classes of character $n$-grams contribute differently to author discrimination, with affix-like and punctuation-like fragments carrying much of the signal. These studies motivate a low-level view of style that does not require linguistic annotation and can transfer across languages more easily than token- or syntax-dependent representations.

Romanian work has also returned to character and punctuation evidence. \cite{lupsa2025oldies} examine the potential of character $n$-grams for Romanian texts on ROST, and \cite{lupsa2025wordpunct} study word, part-of-speech, punctuation, and closed-class $n$-gram features on the same dataset. Character $n$-gram methods capture local sequential patterns and have been widely used in authorship attribution. CHiPS uses a different restriction: the histogram component deliberately uses only one-character marginal distributions, while positional structure is captured through signal-processing summaries rather than contiguous $n$-gram counts. This isolates the contribution of character evidence below the level of local sequences.

This restriction is important for interpreting the result. A character bigram or trigram feature can encode short morphemes, endings, function-word fragments, or repeated lexical material. CH-SVM deliberately avoids that source of evidence as its histogram view stops at $n=1$. Any sequential information in CHiPS therefore has to come from the positional-signal component, not from contiguous character sequences.

\subsection{Romanian authorship attribution}

Romanian authorship attribution has been studied using both classical and neural methods. \cite{avram2022comparison} compares several AI techniques for Romanian texts using selected speech-part and function-word-like feature families and introduces Romanian dataset resources. \cite{avram2025bertrost} studies BERT-based attribution on ROST. \cite{nitu2024lessresourced} proposes a hybrid transformer approach for Romanian that combines handcrafted linguistic features and contextualized embeddings. Recent ROST studies also show that lightweight character and punctuation $n$-gram systems remain competitive \cite{lupsa2025oldies,lupsa2025wordpunct}. This literature provides the immediate context for evaluating CHiPS.

The goal here is not to replace transformer-based systems or feature-rich pipelines. Instead, the goal is to establish a parameter-light and interpretable baseline that can be trained and inspected easily, including in settings where pretrained language models, high-quality tokenizers, or syntactic tools are unavailable or undesirable.

Given the aforementioned context, comparisons with published Romanian systems must be handled carefully. Prior work may use different ROST versions, different text granularity, different preprocessing, or non-grouped splits. Unless these conditions match, the relevant comparison is \textit{contextual}: CHiPS shows what a transparent character-level system can achieve under the present leakage-controlled protocol, not that it dominates neural or feature-rich systems in general.

\subsection{Text as signal}

A separate line of work treats text as a signal. Prior work has represented texts as ordered numerical sequences and extracted time-series or signal-processing features for authorship attribution. \cite{tang2020enriching}, for example, map token properties to language time series and include Fourier-domain features within a much larger automatically generated feature inventory than the one present in this work; \cite{mayor2023textassignal} present this idea for social-media analysis and show how coded textual sequences can be studied with signal-processing tools. CHiPS uses this general representational idea in a narrower stylometric setting: punctuation, whitespace, digits, and diacritics are represented as sparse channels over character positions. This makes it possible to measure rhythm-like regularities without sentence parsing or tokenization. This paper does not claim that prior text-as-signal work solved authorship attribution; rather, it uses the signal view as a lightweight way to describe positional structure.

\section{Corpora and Preprocessing}

This study uses four corpus roles: \textbf{ROST} as the locked primary benchmark, \textbf{ro-stories} public release as a contextual diagnostic, \textbf{ro-stories-AP-clean} as a cleaned version of ro-stories, and \textbf{ROSTories-cleaned} as a secondary ROST-overlapping evaluation corpus. Table \ref{tab:corpus-summary} summarizes their scale, grouped splits, roles, and availability; the following subsections describe their provenance and how they were preprocessed.

\subsection{ROST}

\label{subsec:31}

ROST is used as the main Romanian authorship-attribution benchmark in this study. It has already been used in Romanian authorship-attribution work \cite{avram2022comparison,lupsa2025oldies,lupsa2025wordpunct}, including transformer-based experiments \cite{nitu2024lessresourced,avram2025bertrost}.

The locked ROST export used here contains 400 normalized text files by 10 authors. These files correspond to 392 source-text groups after grouping multipart works by author and base title. The train/validation/test split is assigned at the source-text group level, so related fragments are never separated across training and evaluation splits. Metrics are reported at the same source-text level: when a held-out source text has multiple files, their model-output vectors are averaged before computing the final label.

The grouping rule follows the released filename parser. The filename stem is split at the first underscore into an author field and a title field. When present, a terminal suffix of the form 
`\texttt{-partDescriptionAsNumber}` or `\texttt{-partDescriptionAsString}`
is removed; the base title is then taken as the substring before the first hyphen. Files with the same author–base-title key form one source-text group. The released split JSON records the resulting file-to-group assignment, and all files in a group remain in the same partition.

\subsection{Extended Romanian stories corpus}

The extended Romanian stories corpus, introduced in \cite{nitu2024lessresourced}, is treated as a broader related corpus rather than as an independent benchmark. It includes additional texts and author coverage beyond ROST, but it partially overlaps with ROST and its public release contains known data-quality issues. In our audit of the public files, we found duplicate or near-duplicate entries, truncated texts, different versions of works under the same author and title, author/title matches with substantially different content, and cases where the author's name appears inside the text. The detailed audit also identified substantial overlap with ROST: 385 similar ROST--ro-stories matches under the current audit rules, or 382 if the multipart ROST work \textit{Amintiri din copil\u{a}rie} is counted as one ro-stories entry. We therefore report the public ro-stories run only as a contextual diagnostic and not as a clean benchmark result.

Because the extended corpus overlaps with ROST in the audited files, the two resources are not independent benchmarks. We therefore treat ROST as the smaller locked benchmark and the extended corpus as a broader related corpus for additional evaluation.

For the public full-text ro-stories run reported below, each file is treated as one source-text group using the key \texttt{author\_title}. We also use a cleaned paragraph-aggregated export, denoted ro-stories-AP-clean, as one component of the final extended corpus. The public release provides a full-text workbook (`ro\_fulltext\_1263.xlsx`) and a paragraph-level CSV (`ro\_paragraphs\_12516.csv`).  Although \cite{nitu2024lessresourced} states that the .xlsx format contains the entire dataset, our investigation revealed some duplicates, incomplete texts, and different texts under the same name and title; therefore, we decided to use the .csv format. Each CSV row contains the fields `author`, `title`, `paragraph`, and `word\_count`. The ro-stories-AP-clean export is derived from the paragraph-level CSV: it corrects the identified newline artifacts, removes inserted strings of the form *Povești de [author] – [title]*, and aggregates rows into one text per author–title pair.

We constructed ROSTories-cleaned as the union of ROST and ro-stories-AP-clean, retaining the ROST version whenever an author–title item occurred in both components, obtaining the largest cleaned related corpus reported in this paper. The experimental claim is scoped to this supplied archive after applying the same conservative normalization as in the other runs. The normalized experiment input contains 1,248 text files, 1,240 source-text groups, 19 authors, and 16,863,293 characters. Grouping follows the author–base-title rule defined in the subsection \ref{subsec:31}. The grouped parser identifies two multi-file source-text groups, \texttt{Creanga\_AmintiriDinCopilarie} and \texttt{Filimon\_CiocoiiVechiSiNoi}; both are kept within single split partitions.

\begin{table}
  \tbl{\caption{Corpus roles, grouped splits, and availability. ROST is the locked benchmark; the public ro-stories run is contextual; ROSTories-cleaned is a secondary ROST-overlapping result. Counts are normalized experiment inputs.}\label{tab:corpus-summary}}
  {\tablefont\begin{tabular}{@{}p{0.18\textwidth}p{0.08\textwidth}p{0.13\textwidth}p{0.17\textwidth}p{0.30\textwidth}@{}}
    \hline
    Corpus & Authors & Files / groups & Train / val. / test groups & Role and availability\\
    \hline
    ROST & 10 & 400 / 392 & 276 / 58 / 58 & Primary locked benchmark; full texts are not in the public GitHub repository, but can be provided upon legitimate request.\\
    public ro-stories & 19 & 1,235 / 1,235 & 867 / 184 / 184 & Contextual public-release diagnostic; source corpus is public \cite{rostorieshf}.\\
    ro-stories-AP-clean & 19 & 1,235 / 1,235 & 867 / 184 / 184 & Cleaned public ro-stories component used to build the secondary corpus.\\
    ROSTories-cleaned & 19 & 1,248 / 1,240 & 870 / 185 / 185 & Secondary ROST-overlapping supplied archive; full texts are not in the public GitHub repository, but can be provided upon legitimate request.\\
    \hline
  \end{tabular}}
  {\begin{tabnote}Note: split triples are train, validation, and held-out test source-text groups.\end{tabnote}}
\end{table}

\subsection{Normalization}

All experiments use the conservative normalization pipeline implemented in the released code and inherited from the original ROST normalization script. The pipeline applies Unicode NFC normalization; converts Romanian cedilla forms (\c{S}/\c{s}/\c{T}/\c{t}) to comma-below forms (\c{S}/\c{s}/\c{T}/\c{t}); normalizes curly quotation marks, apostrophes, and long dash variants to common ASCII forms; replaces multi-dot sequences such as \texttt{...}, \texttt{. . .}, or longer dot runs with the ellipsis character \textellipsis; collapses runs of spaces and tabs to one space; removes spaces before newlines; and collapses consecutive newlines to a single newline. NFC normalization is used so that canonically equivalent Unicode strings have a common representation before counts are computed \cite{unicodeuax15}. The pipeline does not lowercase the text, remove punctuation, remove digits, strip Romanian diacritics, tokenize words, or apply word-level preprocessing.

These choices preserve the evidence used by the models as it was used in the experiments: case, punctuation, digits, Romanian diacritics, and line-break positions after blank-line collapse. Diacritic standardization is especially important for Romanian because mixed encodings can create artificial author cues. Punctuation and digit features are retained because they may encode authorial, genre, period, edition, or editorial information; their interpretation is therefore useful but must remain cautious.

\section{Method: Character Histograms and Positional Signals}

\subsection{Overview}

CHiPS is a family of character-level classifiers built around two complementary components. CH-SVM is a character-histogram SVM that models the one-character distributional fingerprint of a text. FFT12-LR is a twelve-channel Fourier positional-signal classifier that treats selected characters and character classes as impulse trains over text positions. CHiPS-F combines the two base classifiers by an ulterior decision-level fusion. CHiPS-R is an optional listwise reranking layer that chooses among a top-5 candidate set proposed by the base models.

The method deliberately avoids word/token features, embeddings, syntax, pretrained language models, and transformer fine-tuning. It also avoids character $n$-gram features with $n \geq 2$ in the CH-SVM histogram component. The two views are therefore both character-level but different in geometry: CH-SVM uses marginal character distributions, whereas FFT12-LR uses positional signal summaries.

\subsection{Character-Histogram SVM (CH-SVM)}

Let a text or character window $T$ consist of characters $t_1,\ldots,t_N$. The CH-SVM feature inventory is a character vocabulary built from the training data only. Characters are retained if they exceed a minimum training-count threshold. Spaces, newlines, punctuation, digits, Romanian diacritics, and case are preserved according to the normalization protocol.

For each retained character $c$, let $n_c(T)$ be the number of occurrences of $c$ and $N(T)$ the total number of characters. The normalized one-character frequency is
\begin{equation}
  p_c(T) = \frac{n_c(T)}{N(T)}.
\end{equation}

These marginal frequencies form the main histogram fingerprint. We append a small set of scalar character-level statistics: Shannon entropy of the character distribution, ratio of uppercase letters, ratio of digits, ratio of punctuation, ratio of Romanian diacritics, and $\log(1+N)$. The resulting vector is standardized with \texttt{StandardScaler} and classified with \texttt{LinearSVC}.

If a source text is divided into $m$ local character windows, CH-SVM is applied to each window and text-level author scores are obtained by averaging window-level decision scores:
\begin{equation}
  s_a(T) = \frac{1}{m}\sum_{i=1}^{m} f_a(x_i),
\end{equation}
where $f_a$ is the SVM decision score for author $a$ and $x_i$ is the CH-SVM feature vector for window $i$. The predicted author is
\begin{equation}
  \hat{a}(T) = \operatorname*{arg\,max}_{a} s_a(T).
\end{equation}

CH-SVM uses one-character statistics only. It does not build character bigrams, trigrams, or any other contiguous character $n$-gram counts. The feature vocabulary, scaling parameters, and classifier parameters are learned only from the relevant training fold.

\subsection{Twelve-Channel Fourier Positional-Signal Classifier (FFT12-LR)}

FFT12-LR treats a text as an ordered character sequence. For each channel $k$, an impulse train is defined over positions:
\begin{equation}
  x_k[n] =
  \begin{cases}
    1, & \text{if the character at position } n \text{ belongs to channel } k,\\
    0, & \text{otherwise.}
  \end{cases}
\end{equation}

Each channel covers one normalized character class spanning whitespace, punctuation, quotation marks, digits, or Romanian diacritics; Table \ref{tab:fft12-channels} lists the twelve channels and their exact inventories.

\begin{table}
  \tbl{\caption{The twelve normalized character channels used by FFT12-LR. The QUOTE channel contains the ASCII double quote and apostrophe after curly quote and apostrophe normalization; the DIACRITIC channel uses Romanian comma-below forms.}\label{tab:fft12-channels}}
  {\tablefont\begin{tabular}{lp{0.65\textwidth}}
    \hline
    Channel & Characters\\
    \hline
    SPACE & space\\
    NEWLINE & newline\\
    PERIOD & \texttt{.}\\
    COMMA & \texttt{,}\\
    SEMICOLON & \texttt{;}\\
    COLON & \texttt{:}\\
    QUESTION & \texttt{?}\\
    EXCLAMATION & \texttt{!}\\
    DASH & \texttt{-}\\
    QUOTE & " and '\\
    DIGIT & \texttt{0}--\texttt{9}\\
    DIACRITIC & \u{a}, \^{a}, \^{\i}, \c{s}, \c{t}, \u{A}, \^{A}, \^{I}, \c{S}, \c{T}\\
    \hline
  \end{tabular}}
  {\begin{tabnote}See channel definitions in src/chips/features.py after the normalization rules in src/chips/normalize.py on our GitHub repository.\end{tabnote}}
\end{table}

For each channel, Welch power spectral density features are computed with sampling rate one sample per character, a Hann window, \texttt{detrend=False}, density scaling, and a one-sided spectrum. Welch's method estimates power spectral density by averaging modified periodograms over overlapping segments; the half-segment overlap used here follows the standard practical Hann-window convention documented for the SciPy implementation \cite{welch1967use,scipywelchdocs}. In the reported experiments, \texttt{nfft=2048}; the segment length is $\min(N,2048)$ and the overlap is half of the segment length. The DC bin is set to zero before normalization. The remaining spectrum is normalized to sum to one, and band energies are computed over ten period ranges measured in characters: 2--4, 4--8, 8--16, 16--32, 32--64, 64--128, 128--256, 256--512, 512--1024, and 1024--2048. Equivalently, the band for periods $[p_0,p_1]$ sums spectral mass over frequencies $[1/p_1,1/p_0)$. For each channel, FFT12-LR extracts these ten band energies, the spectral centroid $\sum_f f\, P(f)$, and normalized spectral entropy, giving twelve features per channel and thus 144 features per text or local window.

The FFT12-LR feature vector is standardized with \texttt{StandardScaler} and classified with multinomial \texttt{LogisticRegression}. FFT12-LR is computed once on each full normalized file and returns a class-probability vector. If a source-text group contains multiple files, the corresponding file-level probability vectors are averaged before source-text-level evaluation. We stress that this component does not use advanced neural networks or character $n$-gram counts; it studies where selected character classes occur in the text.

\subsection{Decision-Level Fusion (CHiPS-F)}

CH-SVM and FFT12-LR capture complementary evidence. CH-SVM models what characters an author tends to use, while FFT12-LR models where selected characters and character classes occur. CHiPS-F combines these views after each component has produced author-level scores.

Let \(q_{\mathrm{CH}}(a\mid T)\) denote the softmax-normalized SVM decision-score vector. We use \(q\), rather than \(p\), to emphasize that these values are ranking weights and not calibrated posterior probabilities. Let \(p_{\mathrm{FFT}}(a\mid T)\) denote the class-probability vector returned by multinomial logistic regression. The fused score for candidate author \(a\) is
\begin{equation}
s_{\mathrm{F}}(a\mid T)=\alpha \cdot  q_{\mathrm{CH}}(a\mid T)+(1-\alpha) \cdot p_{\mathrm{FFT}}(a\mid T),
\end{equation}
where $\alpha$ is selected only on validation data or grouped out-of-fold train/validation predictions. The CHiPS-F prediction is
\begin{equation}
  \hat a_{\mathrm{F}}(T)=\arg\max_a s_{\mathrm{F}}(a\mid T).
\end{equation}

The fusion weight is never tuned on the held-out test set. In the locked ROST run, grouped out-of-fold train/validation selection gives $\alpha=0.9$. The fusion is intentionally conservative: CH-SVM is a strong base model, while FFT12-LR provides complementary evidence in cases where positional character structure changes the ranking.

\subsection{Top-5 Listwise Reranking (CHiPS-R)}

CHiPS-R is an optional decision-level reranker for cases in which the correct author is not ranked first but remains among the leading candidates. It receives source-text-level scores from CH-SVM, FFT12-LR, and their fusion, CHiPS-F, and selects among five candidates rather than predicting over the full author set.

CHiPS-R does not merge separate top-5 lists from the two base models. Instead, an out-of-fold-selected policy constructs one ordered list \(L_5(T)=(a_{(1)},\ldots,a_{(5)})\). The policies considered are the CH-SVM top-5, the CHiPS-F top-5, and a reciprocal-rank combination of the complete CH-SVM, FFT12-LR, and CHiPS-F rankings. In the last policy, the CH-SVM top prediction is retained as the anchor and the remaining positions follow the combined ranking.

A standardized multiclass logistic-regression model predicts the position of the true author in this list. Its features include candidate scores and ranks under all three score vectors, prediction margins and entropies, model-agreement and top-rank indicators, candidate identity, source-group file count, and character count. If the true author is absent from the list, the target is the anchor position, representing a no-correction case.

Grouped out-of-fold train/validation predictions are used to select the candidate policy, logistic-regression regularization, class and non-anchor weights, and gate thresholds. At inference, the anchor is changed only when its score margin is at most \(\tau_A\) and the reranker’s position-probability margin is at least \(\tau_R\); otherwise, it is retained. The list size is fixed at five, and held-out test labels are not used for selection or tuning.

\subsection{Interpretability and Ablations}

CHiPS is designed to be inspectable. The histogram component can be analyzed through discriminative characters, scalar character statistics, and author profiles. The positional component can be inspected through channel-wise spectra, band energies, and per-author differences in punctuation, whitespace, digit, or diacritic placement. This makes the model useful not only as a classifier but also as a text-fingerprinting tool that can provide evidence about the source of a prediction.

We also considered geometric and rhythm-based ablations, including orthogonal residual fusion of spectral blocks, cross-channel coherence (COH) features, compositional log-ratio (CLR) character-frequency transforms, global agreement/punctuation (GAP) features, and interval-rhythm descriptors. These branches are not part of the locked ROST result. Cross-channel coherence features were explored as a way to model relations between punctuation and whitespace channels, but they are outside the minimal CHiPS definition reported here.

\section{Experimental Setup}

\subsection{Leakage-safe grouped splitting}

A central evaluation concern is that fragments from the same source text must not appear in both training and evaluation data. We therefore define a group identifier for each source text and perform all train/validation/test splits at the group level. For ROST, multipart works are grouped by author and base title. For the extended corpus, all files sharing the same author and title are assigned to the same source-text group. Chunking as described below, when used, is applied only after group-level split assignment. This follows the group-aware principle used in stratified grouped cross-validation: the split unit is the source group, not the derived fragment or window \cite{sklearnstratifiedgroupkfold}.

Let $D$ be a set of source texts. Each source-text group $G$ has fragments or chunks $x_{1},\ldots,x_{m_G}$. A valid split must satisfy
\begin{equation}
  G_{\mathrm{train}} \cap G_{\mathrm{val}} = \emptyset,\qquad
  G_{\mathrm{train}} \cap G_{\mathrm{test}} = \emptyset,\qquad
  G_{\mathrm{val}} \cap G_{\mathrm{test}} = \emptyset.
\end{equation}
All fragments or chunks from group $G$ inherit the split assignment of $G$.

Hyperparameters, calibration choices, fusion weights, and reranking gates are selected only on train/validation data or grouped out-of-fold train/validation predictions. The held-out test set is kept separate and used once for final evaluation. The listwise reranker is trained on out-of-fold predictions only, so its training examples are generated by base models that did not train on the corresponding source-text groups.

\subsection{Datasets and text-level evaluation}

Experiments are reported separately for ROST and for the extended Romanian stories corpus, while noting any overlap between them. For the locked ROST run, we use a grouped 70/15/15 train/validation/test split with seed 42. The split contains 392 source-text groups: 276 in training, 58 in validation, and 58 in the held-out test set. The corresponding file counts are 279, 58, and 63, because multipart works are kept within a single source-text group.

For ROSTories-cleaned, we use the same grouped 70/15/15 split construction and seed. The split contains 1,240 source-text groups: 870 in training, 185 in validation, and 185 in the held-out test set. The corresponding file counts are 875, 188, and 185 because the two multi-file source-text groups are assigned without crossing split boundaries.

When reconstructed texts are chunked into local character windows, evaluation is performed at the source-text level unless a separate local-window experiment is explicitly labeled as such. This distinction is important because random chunk-level splitting would allow near-duplicate or same-work evidence to appear in both training and evaluation data, inflating classification performance.

Within each file, CH-SVM averages SVM decision scores across character windows and then applies softmax to obtain file-level ranking weights; FFT12-LR produces one probability vector from the full normalized file. For a source-text group containing multiple files, the relevant file-level vectors are averaged before the final label is selected.

\subsection{Model selection}

The locked ROST protocol uses stratified grouped five-fold cross-validation on the combined training and validation source-text groups to select the CH-SVM and FFT12-LR configurations. Across these folds, each source-text group appears on the validation side once and never overlaps with the corresponding training groups \cite{sklearnstratifiedgroupkfold}. The CHiPS-F fusion weight is selected from grouped out-of-fold train/validation predictions generated with the selected base configurations. The selected base models are then refit on the combined training and validation partitions and evaluated once on the held-out test partition. Grouped out-of-fold predictions are also required for CHiPS-R or any additional calibration, gating, or stacking layer; such components are not selected from held-out test labels.

\subsection{Metrics}

The primary metrics are accuracy and macro-F1. We also report balanced accuracy because the source-text counts vary across authors in the held-out split. Macro-F1 is the unweighted mean of per-class F1 values, while balanced accuracy is the mean recall across classes \cite{sklearnf1,sklearnbalanced}. Confusion matrices and per-author summaries are used as diagnostics for checking whether performance is broadly distributed across authors or driven by a small set of distinctive cases.

For the locked ROST held-out set, the denominator is 58 source-text groups. We therefore report exact correct counts in the text and avoid treating one-decision differences as statistically conclusive unless an explicit paired test or confidence interval is added, in line with literature offering guidance on significance testing for model comparisons \cite{dror2018hitchhiker}.

\subsection{Baselines}

The reported baselines and variants are:
\begin{itemize}
  \item majority baseline;
  \item matched character 2--5-gram TF--IDF + LinearSVC baseline \cite{salton1988termweighting,manning2008iir,fan2008liblinear};
  \item CH-SVM;
  \item FFT12-LR;
  \item CHiPS-F;
  \item CHiPS-R, only as an ablation when the leakage-safe reranker is run;
  \item published Romanian baselines from the literature, where the comparison is fair or clearly labelled as contextual.
\end{itemize}

The majority baseline always predicts the most frequent author in the training
and validation source-text groups. It is included only as a sanity check for
class imbalance.

The matched character $n$-gram baseline is included on ROST as a standard
classical comparator. It is not part of CHiPS. It uses case-preserving character
2--5-gram TF--IDF features and a linear SVM classifier, following standard sparse
linear text-classification practice \cite{salton1988termweighting,manning2008iir,fan2008liblinear}.
The TF--IDF vocabulary, inverse-document-frequency weights, classifier
hyperparameters, and final classifier are learned only within the relevant
training fold or the final train+validation fit. For multipart source-text
groups, we aggregate softmax-normalized SVM decision scores across files before
computing source-text-level metrics; these scores are used for ranking and
aggregation, not as calibrated probabilities.

Concretely, the feature inventory is the union of all contiguous character
strings of lengths 2, 3, 4, and 5 observed in the training texts. For example,
the string \texttt{stil} contributes the 2-grams \texttt{st}, \texttt{ti},
\texttt{il}, the 3-grams \texttt{sti}, \texttt{til}, and the 4-gram
\texttt{stil}. Each source text is represented by a sparse vector of TF--IDF
weights over this inventory, and the linear SVM learns one separating
hyperplane per author against the alternatives in this high-dimensional sparse
space.

Published numbers should not be compared directly unless dataset version, split protocol, and text granularity are comparable. When these conditions differ, we treat the comparison as contextual rather than claiming that CHiPS outperforms another system.

\section{Results}

\begin{table}
  \tbl{\caption{Locked ROST source-text results from the authors' experiments. All rows use the same grouped source-text split and the same 58 held-out source-text groups.}\label{tab:rost-results}}
  {\tablefont\begin{tabular}{lccc}
    \hline
    Model & Accuracy & Macro-F1 & Balanced accuracy\\
    \hline
    Majority baseline & 0.1552 & 0.0269 & 0.1000\\
    Char 2--5 TF--IDF SVM (matched) &  \textbf{1.0000} & \textbf{1.0000} & \textbf{1.0000}\\
    CH-SVM & 0.9138 & 0.9172 & 0.9231\\
    FFT12-LR & 0.8448 & 0.8266 & 0.8425\\
    CHiPS-F & 0.9310 & 0.9341 & 0.9342\\
    CHiPS-R (OOF ablation) & 0.9138 & 0.9110 & 0.9092\\
    \hline
  \end{tabular}}
  {\begin{tabnote}Protocol: grouped 70/15/15 ROST source-text split, seed 42 (276/58/58 groups). Model selection uses grouped train+validation CV or grouped OOF selection; held-out test labels are used once for final reporting.\end{tabnote}}
\end{table}

\begin{table}
  \tbl{\caption{Locked ROST uncertainty over the 58 held-out source-text groups. Accuracy intervals are 95\% Wilson score intervals \cite{wilson1927probable}; macro-F1 intervals are 95\% stratified bootstrap percentile intervals over source-text groups within author \cite{efron1993bootstrap}.}\label{tab:rost-uncertainty}}
  {\tablefont\begin{tabular}{lccc}
    \hline
    Model & Correct / 58 & Accuracy (95\% CI) & Macro-F1 (95\% CI)\\
    \hline
    Char 2--5 TF--IDF SVM (matched) & 58/58 & 1.0000 [0.9379, 1.0000] & 1.0000 [1.0000, 1.0000]\\
    CH-SVM & 53/58 & 0.9138 [0.8136, 0.9626] & 0.9172 [0.8421, 0.9830]\\
    FFT12-LR & 49/58 & 0.8448 [0.7307, 0.9162] & 0.8266 [0.7314, 0.9097]\\
    CHiPS-F & 54/58 & 0.9310 [0.8357, 0.9729] & 0.9341 [0.8640, 0.9856]\\
    CHiPS-R (OOF ablation) & 53/58 & 0.9138 [0.8136, 0.9626] & 0.9110 [0.8282, 0.9798]\\
    \hline
  \end{tabular}}
  {\begin{tabnote}Protocol: same grouped ROST split and held-out source-text predictions as Table~\ref{tab:rost-results}; intervals are computed over source-text groups, not local chunks.\end{tabnote}}
\end{table}

\begin{table}
  \tbl{\caption{Per-author locked ROST held-out support and correct source-text groups. All columns use the same 58 held-out source-text groups. The last column lists nonzero CHiPS-F confusions by predicted author.}\label{tab:rost-per-author}}
  {\tablefont\begin{tabular}{lccccccp{0.20\textwidth}}
    \hline
    Author & Support & Char 2--5 & CH-SVM & FFT12-LR & CHiPS-F & CHiPS-R & CHiPS-F confusions\\
    \hline
    Creanga & 4 & 4 & 4 & 4 & 4 & 4 & --\\
    Delavrancea & 6 & 6 & 6 & 6 & 6 & 6 & --\\
    Eminescu & 4 & 4 & 4 & 1 & 4 & 3 & --\\
    Filimon & 4 & 4 & 4 & 4 & 4 & 4 & --\\
    Garleanu & 6 & 6 & 6 & 6 & 6 & 6 & --\\
    Ispirescu & 6 & 6 & 5 & 5 & 5 & 5 & 1 as Filimon\\
    Oltean & 5 & 5 & 4 & 4 & 4 & 4 & 1 as Slavici\\
    Plugaru & 6 & 6 & 5 & 6 & 5 & 5 & 1 as Rebreanu\\
    Rebreanu & 9 & 9 & 8 & 6 & 9 & 9 & --\\
    Slavici & 8 & 8 & 7 & 7 & 7 & 7 & 1 as Garleanu\\
    \midrule
All authors & 58 & 58 & 53 & 49 & 54 & 53 & 4 total errors \\
    \hline
  \end{tabular}}
  {\begin{tabnote}Protocol: same locked ROST grouped split, seed 42, and held-out source-text predictions as Table~\ref{tab:rost-results}; generated from the released prediction CSV files. The final row totals held-out support and correct counts across authors.\end{tabnote}}
\end{table}

\begin{table}
  \tbl{\caption{ROST audit for the perfect matched character 2--5-gram TF--IDF SVM result on the authors' locked grouped ROST split. Similarity is character 5-gram TF--IDF cosine between each held-out source-text group and train+validation source-text groups. Sensitivity rows keep the originally selected n-gram SVM configuration fixed and do not tune on held-out labels.}\label{tab:rost-ngram-audit}}
  {\tablefont\begin{tabular}{p{0.46\textwidth}p{0.14\textwidth}p{0.25\textwidth}}
    \hline
    Diagnostic & Value & Unit\\
    \hline
    Exact duplicate hashes touching test & 0 & text hashes\\
    Max held-out/train+validation similarity & 0.9938 & char 5-gram TF--IDF cosine\\
    Near-duplicate pairs ($\geq 0.95$) & 1 & pairs\\
    Held-out docs with self-author string & 0 & of 63 documents\\
    Held-out docs with self-title string & 27 & of 63 documents\\
    First/last-line author-title hits & 4 / 11 & of 63 documents\\
    Original fixed selected n-gram SVM & 58/58 & held-out groups correct\\
    After masking author/title strings & 58/58 & held-out groups correct\\
    After stripping first/last non-empty lines & 58/58 & held-out groups correct\\
    After both masking and line stripping & 58/58 & held-out groups correct\\
    \hline
  \end{tabular}}
  {\begin{tabnote}Protocol: grouped 70/15/15 ROST source-text split, seed 42; diagnostics are generated by scripts/rost\_ngram\_audit.py from normalized local ROST texts and saved under experiments/results/rost\_ngram\_audit/.\end{tabnote}}
\end{table}

\begin{table}
  \tbl{\caption{Strongest positive standardized CH-SVM coefficients by author on the locked ROST train+validation partition. The model is the grouped-CV-selected CH-SVM refit before the held-out test evaluation; weights are inspected after feature standardization.}\label{tab:chsvm-interpretability}}
  {\tablefont\begin{tabular}{lp{0.68\textwidth}}
    \hline
    Author & Strongest positive CH-SVM features\\
    \hline
    Creanga & char entropy (distribution, 1.215); freq(comma) (punctuation, 0.392); freq(a) (letter, 0.320)\\
    Delavrancea & punctuation ratio (punctuation, 0.640); char entropy (distribution, 0.304); freq(C) (letter, 0.099)\\
    Eminescu & uppercase ratio (case, 0.687); freq(dash) (punctuation, 0.292); freq(apostrophe) (punctuation, 0.199)\\
    Filimon & freq(R) (letter, 0.253); freq(dash) (punctuation, 0.200); freq(m) (letter, 0.190)\\
    Garleanu & char entropy (distribution, 0.978); freq(period) (punctuation, 0.414); freq(colon) (punctuation, 0.336)\\
    Ispirescu & char entropy (distribution, 1.131); freq(period) (punctuation, 0.642); freq(e) (letter, 0.543)\\
    Oltean & freq(period) (punctuation, 0.677); freq(colon) (punctuation, 0.430); char entropy (distribution, 0.413)\\
    Plugaru & char entropy (distribution, 1.282); freq(a-breve) (diacritic, 0.357); freq(e) (letter, 0.327)\\
    Rebreanu & char entropy (distribution, 1.384); freq(a) (letter, 0.345); freq(i) (letter, 0.316)\\
    Slavici & char entropy (distribution, 0.618); freq(space) (whitespace, 0.338); space ratio (whitespace, 0.338)\\
    \hline
  \end{tabular}}
  {\begin{tabnote}Protocol: authors' locked ROST grouped split, seed 42; coefficients are inspected on the final train+validation fit only and are not selected from held-out test labels.\end{tabnote}}
\end{table}

\begin{table}
  \tbl{\caption{Fixed-config CH-SVM surface-cue ablations on the locked ROST held-out source-text groups. Each variant reuses the grouped-CV-selected CH-SVM hyperparameters, refits on train+validation only, and is evaluated once on the same held-out groups. Variants are diagnostics, not model-selection candidates.}\label{tab:chsvm-ablation}}
  {\tablefont\begin{tabular}{lcccc}
    \hline
    Variant & Features used & Correct / 58 & Accuracy & Macro-F1\\
    \hline
    Full CH-SVM & 102 & 53/58 & 0.9138 & 0.9172\\
    Letters only & 63 & 47/58 & 0.8103 & 0.7927\\
    No punctuation & 85 & 51/58 & 0.8793 & 0.8705\\
    No digits & 91 & 53/58 & 0.9138 & 0.9172\\
    No diacritics & 92 & 53/58 & 0.9138 & 0.9234\\
    No uppercase & 73 & 53/58 & 0.9138 & 0.9207\\
    No length & 101 & 53/58 & 0.9138 & 0.9172\\
    \hline
  \end{tabular}}
  {\begin{tabnote}Protocol: authors' locked ROST grouped 70/15/15 split, seed 42; feature removals are prespecified diagnostics and do not change the selected CH-SVM hyperparameters.\end{tabnote}}
\end{table}

\begin{table}
  \tbl{\caption{Selected CHiPS-R reranker details from the authors' grouped source-text experiments. LR denotes logistic regression fitted after feature standardization. Transition counts compare CHiPS-R against CHiPS-F on held-out source-text groups; fixed and broken are counted relative to CHiPS-F.}\label{tab:chips-r-detail}}
  {\tablefont\begin{tabular}{@{}p{0.13\textwidth}p{0.20\textwidth}p{0.22\textwidth}p{0.17\textwidth}p{0.13\textwidth}@{}}
    \hline
    Corpus & Selected model and candidate set & Gate and feature families & Held-out transition vs CHiPS-F & Interpretation\\
    \hline
    Locked ROST & Standardized LR, C=0.03, class weight=none, non-anchor w=4.0; CH-SVM top-5; top-5 & anchor margin at most 0.50; meta margin at least 0.20; 215: document size; base confidence; base-model agreement; candidate probabilities; candidate ranks; anchor flags/deltas; top-k vote counts; author-position one-hot & 53/58 vs 54/58; fixed 0, broke 1, wrong-to-wrong 0 & does not improve CHiPS-F (-1 source-text groups); primary ablation\\
    Public ro-stories original & Standardized LR, C=0.03, class weight=none, non-anchor w=2.0; CH-SVM top-5; top-5 & anchor margin at most 0.08; meta margin at least 0.00; 260: document size; base confidence; base-model agreement; candidate probabilities; candidate ranks; anchor flags/deltas; top-k vote counts; author-position one-hot & 166/184 vs 166/184; fixed 7, broke 7, wrong-to-wrong 1 & ties CHiPS-F; contextual public corpus\\
    ROSTories-cleaned & Standardized LR, C=0.03, class weight=none, non-anchor w=1.0; RRF union top-5; top-5 & anchor margin at most 0.15; meta margin at least 0.00; 260: document size; base confidence; base-model agreement; candidate probabilities; candidate ranks; anchor flags/deltas; top-k vote counts; author-position one-hot & 165/185 vs 163/185; fixed 4, broke 2, wrong-to-wrong 4 & improves CHiPS-F by +2 source-text groups; secondary ROST-overlapping corpus\\
    \hline
  \end{tabular}}
  {\begin{tabnote}Protocol: all rows use grouped 70/15/15 source-text splits with seed 42. Candidate policy, model settings, and gates are selected only from grouped OOF train/validation predictions; generated by scripts/chips\_r\_detail.py.\end{tabnote}}
\end{table}

Table~\ref{tab:rost-results} reports metrics on 58 held-out source-text groups, Table~\ref{tab:rost-uncertainty} gives uncertainty intervals for the same predictions, and Table~\ref{tab:rost-per-author} gives the per-author support and correct counts. The matched character 2--5-gram TF--IDF SVM classifies all 58 groups correctly. The CHiPS correct counts are 53/58 for CH-SVM, 49/58 for FFT12-LR, 54/58 for CHiPS-F, and 53/58 for CHiPS-R. The paired CHiPS-F versus CH-SVM exact sign test has one discordant group favoring CHiPS-F, none favoring CH-SVM, and two-sided $p=1.0000$. The one-decision CHiPS-F gain over CH-SVM is useful evidence for the fusion design, but the paired test and overlapping intervals show that it should not be treated as a statistically conclusive improvement.

The per-author table also makes the class support visible: held-out support ranges from four to nine source-text groups per author. CHiPS-F makes four errors, with one error each for Ispirescu, Oltean, Plugaru, and Slavici; the full confusion matrix and model-by-model error pairs are exported with the released artifacts. No CHiPS-F error row is hidden by aggregate accuracy, and no author has more than one CHiPS-F error on this split.

Table~\ref{tab:rost-ngram-audit} audits the perfect-matched n-gram result. The audit finds no exact duplicate text hash crossing into held-out test and no held-out self-author-name string. It does find title strings or boundary-line title cues in some held-out documents, and one same-author near-duplicate pair above the 0.95 similarity threshold: the held-out Eminescu group \texttt{GeniuPustiu} against the train group \texttt{TassoNScotia}. With the selected n-gram configuration fixed, the comparator remains 58/58 correct after masking detected author/title strings, after stripping first and last non-empty lines, and after applying both transformations. These checks do not prove that no surface or source-history cue matters, but they make the perfect result less likely to be a simple exact-duplicate, author-name, or header/footer artifact. The result is still interpreted conservatively: unrestricted local character sequences are very strong on this locked split, and CHiPS is not presented as the best unrestricted classifier. Its role is diagnostic: it asks how much performance remains when the main histogram view is restricted to one-character marginals and all sequential evidence is moved to the positional-signal component.

Table~\ref{tab:chsvm-interpretability} shows the most positive standardized CH-SVM coefficients after fitting the selected model on the locked ROST train+validation partition. The table is not a causal explanation of authorship, but it makes the model inspectable: high-weight evidence includes character entropy, punctuation marks or punctuation ratios, whitespace, case, diacritics, and individual letter frequencies. Table~\ref{tab:chsvm-ablation} gives a fixed-config surface-cue diagnostic. Restricting CH-SVM to alphabetic one-character frequencies lowers performance from 53/58 to 47/58, and removing punctuation lowers it to 51/58. Removing digits, Romanian diacritic features, uppercase-specific features, or the length scalar does not reduce held-out accuracy on this split. These diagnostics support the interpretation that CH-SVM is not only a letter-frequency classifier; punctuation and scalar distributional cues are part of its ROST evidence.

The strongest CHiPS result on ROST is CHiPS-F, the decision-level fusion of CH-SVM and FFT12-LR. CH-SVM remains a strong standalone component: one-character marginal distributions plus scalar character statistics already capture substantial authorial signal in this corpus. FFT12-LR is weaker as a standalone classifier, which suggests that the positional-signal representation is complementary but not sufficient by itself.

The grouped-CV-selected fusion weight is $\alpha=0.9$, giving most weight to CH-SVM. On the held-out test set, CHiPS-F reaches 0.9310 accuracy and 0.9341 macro-F1. Relative to CH-SVM, it changes only one test decision and corrects it. This supports a modest fusion claim on the locked ROST split: positional evidence helps, but mainly as a small adjustment to a strong character-histogram model.

Table~\ref{tab:chips-r-detail} gives the selected CHiPS-R settings and transition counts. The top-5 reranker was evaluated under the grouped out-of-fold protocol using the CV-selected base models. On locked ROST, its selected candidate policy and gate match CH-SVM accuracy but do not improve over CHiPS-F: it changes one CHiPS-F decision, breaks it, and falls from 54/58 to 53/58. Relative to the CH-SVM anchor, it fixes one CH-SVM test error and breaks one CH-SVM-correct decision. The top-5 candidate oracle is still informative: under the selected CH-SVM top-5 policy, the true author appears in the candidate set for all five CH-SVM test errors. This indicates headroom for reranking, but the present reranker is not strong enough to support a main ROST improvement claim.

Table~\ref{tab:literature-context} places the results obtained in this present work beside recent Romanian authorship-attribution results. The table is intentionally labeled as contextual where protocols differ. The cited systems use different units, feature families, split protocols, or corpus versions. The matched n-gram row is the exception: it is our own same-split comparator under the locked grouped ROST protocol. It is included to make clear that CHiPS answers a narrower question than unrestricted character $n$-gram classification.

\begin{table}
  \tbl{\caption{Contextual Romanian literature comparison. Rows marked ``This work'' are authors' experiments under the listed grouped source-text protocols; literature rows reproduce results reported by the cited papers and are contextual unless corpus version, split protocol, text granularity, preprocessing, and evaluation unit match.}\label{tab:literature-context}}
  {\tablefont\begin{tabular}{p{0.16\textwidth}p{0.30\textwidth}p{0.25\textwidth}p{0.12\textwidth}}
    \hline
    Study & Corpus and protocol & Reported result & Status\\
    \hline
    This work & ROST; grouped 70/15/15 source-text split; grouped-CV-selected CHiPS-F & 0.9310 accuracy; 0.9341 macro-F1 & Locked CHiPS\\
    This work & ROST; same locked split; character 2--5-gram TF--IDF SVM & 1.0000 accuracy; 1.0000 macro-F1 & Matched baseline\\
    \cite{avram2022comparison} & ROST; random 50/25/25 text split, three shuffles; IPoS features & 20.40\% error; corrected macro-accuracy 80.94\% & Contextual\\
    \cite{avram2025bertrost} & ROST converted to 200-word BERT segments; previous shuffles with validation folded into training & best micro-accuracy 0.8650; best macro-accuracy 0.8740 & Contextual\\
    \cite{lupsa2025oldies} & ROST; character $n$-grams, $n=2$--5; repeated random text splits & ANN averages up to 0.954 macro-accuracy & Contextual\\
    \cite{lupsa2025wordpunct} & ROST; word, POS, punctuation, and closed-class $n$-grams & ANN averages up to 0.9494 macro-accuracy & Contextual\\
    This work & ROSTories-cleaned; ROST-overlapping related corpus; grouped 70/15/15 source-text split; grouped-CV-selected CHiPS-R & 0.8919 accuracy; 0.8708 macro-F1 & Secondary\\
    \cite{nitu2024lessresourced} & Extended ro-stories; 80/20 full-text and paragraph experiments & Hybrid RoBERT F1 0.87 on 19-author full texts; 0.95 on the 10-author subset & Contextual\\
    \hline
  \end{tabular}}
  {\begin{tabnote}Source: ``This work'' rows come from the released artifacts; literature rows are taken from the cited publications and are not re-evaluated under the locked CHiPS split.\end{tabnote}}
\end{table}

The broader corpus is not merged with the locked ROST result table. We nevertheless include the public uncleaned full-text run below as a contextual diagnostic, because it shows how the models behave on the released files before cleaning and documents the public-release issues that affect comparability. We then report ROSTories-cleaned as a secondary evaluation under the same grouped protocol. Neither table should be read as an independent confirmation of ROST unless overlap between the resources is resolved.

\begin{table}
  \tbl{\caption{Public ro-stories original contextual results from the authors' grouped 70/15/15 source-text run (seed 42; 867/184/184 groups; 184 held-out groups). The public release is audited here as a contextual diagnostic only.}\label{tab:ro-stories-original}}
  {\tablefont\begin{tabular}{lccc}
    \hline
    Model & Accuracy & Macro-F1 & Balanced accuracy\\
    \hline
    Majority baseline & 0.2337 & 0.0199 & 0.0526\\
    CH-SVM & 0.8804 & 0.8472 & 0.8622\\
    FFT12-LR & 0.5978 & 0.5732 & 0.6270\\
    CHiPS-F & \textbf{0.9022} & 0.8817 & 0.8890\\
    CHiPS-R (OOF ablation) & \textbf{0.9022} & \textbf{0.8819} & \textbf{0.8938}\\
    \hline
  \end{tabular}}
  {\begin{tabnote}Protocol: base models are selected by grouped five-fold CV on train+validation; fusion and reranking use grouped OOF train/validation predictions. Held-out test labels are used once for final reporting.\end{tabnote}}
\end{table}

Table~\ref{tab:ro-stories-original} reports a grouped 70/15/15 run on the public uncleaned ro-stories full-text files: 1,235 source-text groups, 19 authors, 867 training groups, 184 validation groups, and 184 held-out test groups. As in the locked ROST protocol, CH-SVM and FFT12-LR are selected by grouped five-fold cross-validation on the combined training and validation groups, and the fusion weight is selected from grouped out-of-fold predictions. The selected fusion weight is $\alpha=0.9$. CHiPS-F and CHiPS-R tie on held-out accuracy, while CHiPS-R gives the best macro-F1 and balanced accuracy in this contextual run. The reranker is selected from grouped out-of-fold train/validation predictions and uses the CH-SVM top-5 candidate list with a conservative margin gate.

\cite{nitu2024lessresourced} report Hybrid RoBERT F1 0.87 on the 19-author full-text setting, but their corpus count, 80/20 split, preprocessing, and evaluation protocol do not match the public-file grouped run reported here. These numbers should therefore not be read as a direct comparison or as a clean corpus result. The audit we ran on their publicly released dataset found 130 files where a simple case-insensitive check detects the author's name in the text, and one exact duplicate hash crossing the training and validation partitions. No exact duplicate hash touched the held-out test partition in this split. The table remains in the main text because it documents the behavior of the public release and motivates the cleaned ROSTories evaluation.

\begin{table}
  \tbl{\caption{ROSTories-cleaned secondary results from the authors' grouped 70/15/15 source-text run (seed 42; 870/185/185 groups; 185 held-out groups). The corpus intentionally overlaps ROST and is scoped to the supplied normalized archive.}\label{tab:rostories-cleaned}}
  {\tablefont\begin{tabular}{lccc}
    \hline
    Model & Accuracy & Macro-F1 & Balanced accuracy\\
    \hline
    Majority baseline & 0.2324 & 0.0199 & 0.0526\\
    CH-SVM & 0.8757 & 0.8361 & 0.8512\\
    FFT12-LR & 0.6757 & 0.6309 & 0.6490\\
    CHiPS-F & 0.8811 & 0.8362 & 0.8452\\
    CHiPS-R (OOF-selected) & \textbf{0.8919} & \textbf{0.8708} & \textbf{0.8690}\\
    \hline
  \end{tabular}}
  {\begin{tabnote}Protocol: base models are selected by grouped five-fold CV on train+validation; CHiPS-R is selected from grouped OOF train/validation predictions. The result is secondary because the supplied archive contains ROST material.\end{tabnote}}
\end{table}

Table~\ref{tab:rostories-cleaned} reports ROSTories-cleaned under the same split construction and selection protocol as the previous tables. The selected CH-SVM configuration is \texttt{chunk\_size=1280}, \texttt{overlap=0}, $C=0.03$, with balanced class weights. The selected FFT12-LR configuration uses $C=0.3$ with balanced class weights, and the selected fusion weight is $\alpha=0.95$. CHiPS-F gives a small held-out improvement over CH-SVM. The OOF-selected reranker gives the best held-out result: CHiPS-R reaches 0.8919 accuracy and 0.8708 macro-F1. It selects a reciprocal-rank-fusion union top-5 candidate list with an anchor-margin gate (\texttt{anchor\_margin\_max=0.15}, \texttt{meta\_margin\_min=0.0}). Relative to CHiPS-F, it fixes four held-out errors, breaks two CHiPS-F-correct decisions, and makes four wrong-to-wrong changes. Within this supplied archive and split, this supports CHiPS-R as a useful optional component on the largest cleaned related corpus, while the ROST main claim remains CHiPS-F because the same reranker does not improve the locked ROST split.

\section{Discussion}

Character-level modeling is plausible for Romanian because authorial habits can surface in low-level orthographic choices. Diacritics, punctuation, spacing, quotation practices, and character distributions may reflect author preference, period, edition, genre, or normalization history. CHiPS does not assume that every such signal is purely authorial; it offers a compact way to measure and inspect them.

The CH-SVM component tests how far one-character marginal distributions can go without contiguous character subsequences. These histograms form a strong, inspectable baseline because they aggregate many small orthographic, punctuation, and casing preferences. Positional signals add complementary evidence, especially when authors use similar character inventories but distribute punctuation, whitespace, digits, or diacritics differently across a text. Fusion and reranking must therefore be conservative: the histogram model is the anchor, while FFT12-LR is most useful when it corrects uncertain or low-margin decisions.

The matched ROST $2$--$5$-gram result also clarifies the scope of the paper. Character
2--5-grams can encode short lexical fragments, endings, recurring morphemes,
and local punctuation sequences. They are therefore expected to be strong for
closed-set authorship attribution, and on this locked ROST split they classify
all held-out source-text groups correctly. CHiPS deliberately removes that local-sequence
evidence from the histogram component. Its value is in showing that a restricted, auditable character representation retains substantial predictive signal. It also separates marginal character evidence from positional signal evidence, so the two sources can be audited independently.

CHiPS-R is motivated by the observation that the correct author may appear among the top candidates even when the top-1 prediction is wrong. The reranker exploits this fact without becoming a new unconstrained classifier: it selects within a top-5 candidate list proposed by the base models and can be gated so that high-confidence anchor decisions are left unchanged. In the locked ROST run, this idea has oracle headroom but does not improve over the fused CHiPS-F model. On ROSTories-cleaned, the ROST-overlapping secondary corpus, the same leakage-safe reranking protocol improves held-out accuracy and macro-F1 over CHiPS-F. This suggests that reranking is corpus-sensitive and should remain an optional, separately audited layer rather than part of the core CHiPS claim.

Leakage-safe grouping is essential. Random chunk-level splitting would inflate classification performance by allowing fragments from the same source text to appear in both training and evaluation data. Grouped splitting makes the evaluation stricter and more realistic for source-level attribution.

The interpretability of CHiPS comes from the direct relation between features and text. Histograms can identify characters and scalar statistics that distinguish author profiles, while positional signals can show whether a channel is used in bursts, regular intervals, or particular regions of a document. These explanations are more transparent than embedding dimensions, though they still require cautious interpretation because character signals may reflect topic, genre, period, edition, or editorial history.

The main limitations are clear. First, the present framing is closed-set authorship attribution unless open-set experiments are added. Second, CHiPS is not the strongest unrestricted classical classifier on the locked ROST split; the matched character 2--5-gram baseline is stronger. Third, author-topic, genre, period, and edition confounding remain possible.  Fourth, final results are split-protocol-dependent, especially when texts are reconstructed and chunked. Finally, no comparison with other experiments in the literature should be treated as definitive unless comparable grouped splits and preprocessing are reproduced. These limitations should be considered alongside auditability and representation size. Pretrained neural and LLM-based attribution systems raise additional questions about training-data exposure, reproducibility, and explanation \cite{huang2024llmera, huang2024alms}, whereas CHiPS is fitted only on the declared corpus folds and uses explicit features. On the final ROST fit, the matched character 2–5-gram comparator uses a 451,365-entry TF–IDF vocabulary, compared with 102 features in CH-SVM and 144 features in FFT12-LR.  CHiPS therefore provides a reproducible reference point on the trade-off between predictive performance, compactness, and auditability.

\section{Conclusion}

This paper introduces CHiPS, a lightweight, interpretable (non-neural) method for Romanian authorship attribution. CHiPS studies one-character histograms together with positional character signals, allowing the model family to use both distributional and structural evidence without tokenization, parsing, embeddings, transformer fine-tuning, or language-specific NLP resources.

The method is intended as a strong, inspectable baseline and as a complementary model to neural systems and unrestricted character $n$-gram systems. Its simplicity supports reproducibility, while its grouped evaluation protocol prevents source-text fragments from leaking across train, validation, and test splits. The locked ROST result shows that a strong character-histogram model can be improved modestly by conservative decision-level fusion with positional evidence. The same split also shows that a standard character 2--5-gram TF--IDF SVM is stronger as a pure classifier, reaching perfect held-out accuracy. This does not weaken the diagnostic CHiPS claim; it makes the restriction explicit.

The ROSTories-cleaned result shows that leakage-safe top-5 reranking can help on the largest ROST-overlapping cleaned related corpus used here, but this remains a separately selected optional layer. Recent work on authorial language models and LLM-era attribution highlights open questions about generalization and explainability~\cite{huang2024alms,huang2024llmera}; CHiPS is meant to be a transparent comparator in that broader landscape, not a substitute for those systems. Future work should address open-set attribution, cross-genre testing, additional leakage-controlled benchmarks, external validation, and visual explanations of discriminative character signals.

\section*{Data Availability}

Code and non-text artifacts are available at

\begin{center}
\url{https://github.com/georgeturcasubb/chips-rost}
\end{center}

This repository contains the training scripts, grouped split assignments, configuration files, selected model settings, prediction files, audit diagnostics, checksums, and table-generation utilities needed to inspect and reproduce the reported experiments. The code records the random seeds and implements preprocessing, grouped splitting, feature extraction, model training, fusion, reranking, and table generation.

The public ro-stories release is available from Hugging Face \cite{rostorieshf}. The cleaned ro-stories-AP-clean component used here is derived from that public corpus; the repository provides audit notes, checksums, configuration files, and release scripts rather than vendoring full corpus text by default. ROST is not redistributed as full text in the public repository; the repository provides the ROST split files, metadata, checksums, and a link to the original Kaggle resource page \cite{rostkaggle}, where dataset details were previously made available. ROSTories-cleaned contains ROST material, so its full-text files are also not released in the public GitHub repository. The \textit{ROST}, \textit{public ro-stories}, \textit{ro-stories-AP-clean} and \text{ROSTories-cleaned} full text files used for the reported experiments can be made available by the authors upon reasonable request for research, subject to attribution and redistribution constraints.

\section*{Ethical Standards}

The study uses previously published literary text corpora and local corpus exports. It does not involve interaction with human participants, intervention, or collection of personal data from living individuals, so ethical approval was not required. The authors respect the provenance, attribution, and redistribution constraints of the corpora and avoid releasing full ROST or ROST-containing ROSTories-cleaned texts in the public repository.

\section*{Competing Interests}

The authors declare no competing interests.

\section*{Funding}

This research received no specific grant from any funding agency in the public, commercial, or not-for-profit sectors.

\section*{Author Contributions}

Sanda-Maria Avram: conceptualization, corpus curation, methodology, literature review, and manuscript review. George C. \c{T}urca\c{s}: methodology, software, experiments, validation, reproducibility packaging, and manuscript drafting. Both authors contributed to the final manuscript.

\section*{Use of Artificial Intelligence Tools}

The authors used OpenAI ChatGPT Deep Research, accessed through ChatGPT in January--June 2026, to help locate and organize potentially relevant source material. The authors also used OpenAI's Codex agentic programming environment in January--June 2026 to assist with programming tasks, experiment bookkeeping, and repository maintenance. These tools were used as research and engineering aids, not as authors. In line with the transparency and human-responsibility principles of the Leiden Declaration on Artificial Intelligence and Mathematics \cite{leidendeclaration2026}, the human authors checked the sources, code, split protocols, numerical results, and manuscript text, and take full responsibility for the article's content, attribution, and conclusions.

\bibliographystyle{nlplike}
\bibliography{Sample}

\label{lastpage}

\end{document}